\documentclass[twoside,twocolumn,9pt]{article}
\usepackage{extsizes}
\usepackage[super,sort&compress,comma]{natbib} 
\usepackage[version=3]{mhchem}
\usepackage[left=1.5cm, right=1.5cm, top=1.785cm, bottom=2.0cm]{geometry}
\usepackage{balance}
\usepackage{mathptmx}
\usepackage{sectsty}
\usepackage{graphicx} 
\usepackage{lastpage}
\usepackage[format=plain,justification=justified,singlelinecheck=false,font={stretch=1.125,small,sf},labelfont=bf,labelsep=space]{caption}
\usepackage{float}
\usepackage{fancyhdr}
\usepackage{fnpos}
\usepackage[english]{babel}
\addto{\captionsenglish}{%
  
}
\usepackage{array}
\usepackage{droidsans}
\usepackage{charter}
\usepackage[T1]{fontenc}
\usepackage[usenames,dvipsnames]{xcolor}
\usepackage{setspace}
\usepackage[compact]{titlesec}
\usepackage{hyperref}
\usepackage{epstopdf}%This line makes .eps figures into .pdf - please comment out if not required.

\definecolor{cream}{RGB}{222,217,201}

\begin{document}

\pagestyle{fancy}
\thispagestyle{plain}
\fancypagestyle{plain}{
%%%HEADER%%%
\renewcommand{\headrulewidth}{0pt}
}
%%%END OF HEADER%%%

%%%PAGE SETUP - Please do not change any commands within this section%%%
\makeFNbottom
\makeatletter
\renewcommand\LARGE{\@setfontsize\LARGE{15pt}{17}}
\renewcommand\Large{\@setfontsize\Large{12pt}{14}}
\renewcommand\large{\@setfontsize\large{10pt}{12}}
\renewcommand\footnotesize{\@setfontsize\footnotesize{7pt}{10}}
\renewcommand\scriptsize{\@setfontsize\scriptsize{7pt}{7}}
\makeatother

\renewcommand{\thefootnote}{\fnsymbol{footnote}}
\renewcommand\footnoterule{\vspace*{1pt}% 
\color{cream}\hrule width 3.5in height 0.4pt \color{black} \vspace*{5pt}} 
\setcounter{secnumdepth}{5}

\makeatletter 
\renewcommand\@biblabel[1]{#1}            
\renewcommand\@makefntext[1]% 
{\noindent\makebox[0pt][r]{\@thefnmark\,}#1}
\makeatother 
\renewcommand{\figurename}{\small{Fig.}~}
\sectionfont{\sffamily\Large}
\subsectionfont{\normalsize}
\subsubsectionfont{\bf}
\setstretch{1.125} %In particular, please do not alter this line.
\setlength{\skip\footins}{0.8cm}
\setlength{\footnotesep}{0.25cm}
\setlength{\jot}{10pt}
\titlespacing*{\section}{0pt}{4pt}{4pt}
\titlespacing*{\subsection}{0pt}{15pt}{1pt}
%%%END OF PAGE SETUP%%%

%%%FOOTER%%%
\fancyfoot{}
\fancyfoot[LO,RE]{\vspace{-7.1pt}\includegraphics[height=9pt]{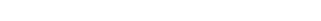}}
\fancyfoot[CO]{\vspace{-7.1pt}\hspace{13.2cm}\includegraphics{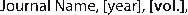}}
\fancyfoot[CE]{\vspace{-7.2pt}\hspace{-14.2cm}\includegraphics{head_foot/RF}}
\fancyfoot[RO]{\footnotesize{\sffamily{1--\pageref{LastPage} ~\textbar  \hspace{2pt}\thepage}}}
\fancyfoot[LE]{\footnotesize{\sffamily{\thepage~\textbar\hspace{3.45cm} 1--\pageref{LastPage}}}}
\fancyhead{}
\renewcommand{\headrulewidth}{0pt} 
\renewcommand{\footrulewidth}{0pt}
\setlength{\arrayrulewidth}{1pt}
\setlength{\columnsep}{6.5mm}
\setlength\bibsep{1pt}
%%%END OF FOOTER%%%

%%%FIGURE SETUP - please do not change any commands within this section%%%
\makeatletter 
\newlength{\figrulesep} 
\setlength{\figrulesep}{0.5\textfloatsep} 

\newcommand{\topfigrule}{\vspace*{-1pt}% 
\noindent{\color{cream}\rule[-\figrulesep]{\columnwidth}{1.5pt}} }

\newcommand{\botfigrule}{\vspace*{-2pt}% 
\noindent{\color{cream}\rule[\figrulesep]{\columnwidth}{1.5pt}} }

\newcommand{\dblfigrule}{\vspace*{-1pt}% 
\noindent{\color{cream}\rule[-\figrulesep]{\textwidth}{1.5pt}} }

\makeatother
%%%END OF FIGURE SETUP%%%

%%%TITLE AND AUTHORS%%%
\twocolumn[
 % \begin{@twocolumnfalse}
%{\includegraphics[height=55pt]{head_foot/journal_name.eps}\hfill\raisebox{0pt}[0pt]
%[0pt]{\includegraphics[height=75pt]{head_foot/RSC_Logo.pdf}}\\[1ex]
%\includegraphics[width=18.5cm]{head_foot/header_bar}}\par
\vspace{1em}
\sffamily
\begin{tabular}{m{4.5cm} p{13.5cm} }

\includegraphics{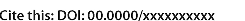} & \noindent\LARGE{\textbf{Designing a Magnetic Micro-Robot for Transporting Filamentous Microcargo}} \\%Article title goes here instead of the text "This is the title"
 & \vspace{0.3cm} \\

 & \noindent\large{Sepehr Ghadami,\textit{${^a}$} Henry Shum,\textit{${^a}$}} \\%Author names go here instead of "Full name", etc.

\includegraphics{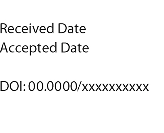} & \\

\end{tabular}

 %\end{@twocolumnfalse} \vspace{0.6cm}
]
%%%END OF TITLE AND AUTHORS%%%

%%%FONT SETUP - please do not change any commands within this section
\renewcommand*\rmdefault{bch}\normalfont\upshape
\rmfamily
\section*{}
\vspace{-1cm}

%%%FOOTNOTES%%%

\footnotetext{\textit{$^{a}$~Department of Applied Mathematics, University of Waterloo ; E-mail: henry.shum@uwaterloo.ca}}

%Please use \dag to cite the ESI in the main text of the article.
%If you article does not have ESI please remove the the \dag symbol from the title and the footnotetext below.

%%%% Abstract text to be placed here %%%%%%%%%%%%

In recent years, the medical industry has witnessed a growing interest in minimally invasive procedures, with magnetic micro-robots emerging as a promising approach. These micro-robots possess the ability to navigate through various media, including viscoelastic and non-Newtonian fluids, enabling targeted drug delivery and medical interventions. Many current designs, inspired by micro-swimmers in biological systems like bacteria and sperm, employ a contact-based method for transporting a payload. Adhesion between the cargo and the carrier can make release at the target site problematic. In this project, our primary objective was to explore the potential of a helical micro-robot for non-contact drug or cargo delivery. We conducted a comprehensive study on the shape and geometrical parameters of the helical micro-robot, specifically focusing on its capability to transport passive filaments. Based on our analysis, we propose a novel design consisting of three sections with alternating handedness, including two pulling and one pushing microhelices, to enhance the capture and transport of passive filaments in Newtonian fluids using a non-contact approach. We then simulated the process of capturing and transporting the passive filament, and tested the functionality of the newly designed micro-robot. Our findings offer valuable insights into the physics of helical micro-robots and their potential for medical procedures and drug delivery. Furthermore, the proposed non-contact method for delivering filamentous cargo could lead to the development of more efficient and effective microrobots for medical applications.
\section{Introduction}
Micro-robots are becoming increasingly popular for various biomedical applications, including drug delivery\cite{drugdelivery1,drugdelivery2,drugdelivery3,drugdelivery4,drugdelivery5,drugDelivery6,drugDelivery7,drugDelivery8}, surgery\cite{surgery1,surgery2,surgery3}, bio-sensing\cite{biosensing}, and chemical and radiation therapies\cite{therapy}. Compared to traditional methods, the use of micro-robots in surgery is less invasive, and they can improve targeted drug delivery while reducing the risk of infection and side effects\cite{surgery1,surgery2,surgery3}. Micro-robots are classified into three groups based on the driving force used for their propulsion and manipulation: chemical\cite{chemicaldriven1,chemicaldriven2}, physical\cite{physicaldriven1,physicaldriven2,physicaldriven3,physicaldriven4}, and bio-hybrid\cite{biohybriddriven1,bioinspired4}. Chemical micro-robots use chemical reactions to generate the desired driving force, but the safety risk of releasing chemical material in vivo limits their usage in such environments. Bio-hybrid micro-robots, which use microorganisms, also pose the risk of leaving microorganisms inside the body. Therefore, physical micro-robots, such as magnetic micro-robots, are more suitable for sensitive environments like inside the body\cite{magneticmicrorobot1,magneticmicrorobot2,magneticmicrorobot3,magneticmicrorobot4}.

Among different types of physical micro-robots, magnetic micro-robots controlled by an external magnetic field are very promising in terms of controllability and safety due to their minimal interaction with tissue. They are capable of penetrating different tissues and have been used in vitreous\cite{magneticmicrorobot1}, knee tissue, and blood vessels\cite{microrobotvessels1,microrobotvessels2}. Magnetic micro-robots are actuated with either torques\cite{magnetictorquedriven} or forces\cite{magneticforcedriven}. In force actuated magnetic micro-robots, a gradient of magnetic field propels the micro-robot. In torque actuated micro-robots, rotating, oscillating or stepping time-varying magnetic fields are applied to propel the micro-robot forward. One of the first helical magnetic micro-robots was designed and fabricated by Bell et al.\cite{firstmicrorobot} The micro-robot was actuated by rotating magnetic fields. When a helical object is rotated about its axis, the drag force applied to the object is asymmetric, which can generate a net driving force to the micro-robot and propel it forward in a direction dependent on the handedness of the helix and the direction of rotation.\cite{purcelHelicalMicrorobot,MicrorobotNew}

Helical magnetic micro-robots have been used for cargo delivery and passive cells. Helical micro-robots generally transport and manipulate cargo using either a direct contact pushing method or a non-contact method. The direct contact pushing method can cause adhesion between the micro-robot and cargo particles, making it challenging to release the cargo when they reach the target.\cite{magneticmicrorobot3} This method can also exert high stress on the cargo and cause mechanical deformations. In contrast, the non-contact method uses the flow generated around the micro-robot to transport it to the target. In 2014, Huang et al. designed a tubular head at the tip of the helical micro-robot to trap cargo by generating microvertices.\cite{noncontactmethod} However, this method was only demonstrated with spherical cargo and may not be effective for transporting long, thin filaments.

One specific application of magnetic micro-robots that has already been demonstrated experimentally is for transporting sperm that are viable but immotile.\cite{magneticmicrorobot3} The dimensions of the helical micro-robot were such that the flagellum of the sperm could be inserted into the centre of the helix but the body of the sperm would not fit through. This trapped the sperm in place and allowed successful transport but release was not always successful as there was adhesion between the micro-robot and the sperm despite attempts to prevent adhesion by functionalizing the surfaces. 

In this study, our objective is to design a magnetic helical micro-robot that utilizes a non-contact method to transport passive filaments along the axis of the robot. The motivation for this is to carry immotile sperm for assisted fertilization, but the design could also be utilized for other filamentous cargo. We begin by describing the model and numerical methods. We then examine the effects of helical geometry on the pathline of a point initially on the axis of a helical micro-robot that rotates without translation. We also study the velocity field around the micro-robot when it is free to translate and rotate subject to a prescribed rotating magnetic field. After presenting these basic properties of the system for a simple helical micro-robot, we show that the micro-robot always moves faster than a filament placed along its axis; hence, the filament cannot be transported effectively with this design. We propose modifying the micro-robot design to include multiple helical sections of alternating handedness. 

In particular, we show that micro-robots constructed from two pulling microhelices and a pushing microhelix in the middle are effective at capturing and transporting the passive filament. Once an intended forward direction for the motion of the micro-robot has been chosen, we define a pulling microhelix to be one that pulls fluid in through the front of the helix, or equivalently pulls the micro-robot forward relative to the fluid. This could either be a right-handed helix rotating about the forward direction or a left-handed helix rotating about the backward direction. A pushing microhelix is one that pushes fluid in the forward direction relative to the micro-robot; this is either a left-handed helix rotating about the forward direction or a right-handed helix rotating about the backward direction. With our design, as the micro-robot approaches a filament, the front pulling helix pulls the filament into the interior of the micro-robot. The middle section pushes the filament forward so that it moves with the same speed as the micro-robot, which is driven primarily by the pulling section at the back. Since pushing and pulling sections oppose each other, the ratio of their lengths determines the overall speed and flow field around the micro-robot. To further investigate the feasibility of fully coupling the helical micro-robot and the filament with this design, we construct a simplified analytical model. Using this model, supplemented with numerically calculated coefficients, we determine the ratios of the pulling and pushing sections of the helix that attain full coupling over a wide range of helical pitches and cross-sectional radii.

\section{Numerical methods and governing equation}

In this project, we develop a simulation model for a passive filament and a helical micro-robot actuated by a magnetic field. For the intended scales associated with the problem, the Reynold's number is small, so we neglect the inertia term in the Navier--Stokes equation and solve the Stokes equation. The governing equations for the incompressible Stokes equation are:

\begin{align}
\label{1 and 2}
\vec{0} = -\nabla p  + \mu\Delta \vec{u} + \vec{F},\\
\nabla \cdot{\vec{u}} = 0,
\end{align}
where $p$ is the pressure, $\vec{u}$ and $\vec{F}$ represent the fluid velocity field and the body forces acting on the fluid, and $\mu$ is the fluid viscosity. Based on our objective of carrying a filament of length 50~\textmu{}m (corresponding to the length of a human sperm flagellum, for example) suspended in an aqueous medium, we non-dimensionalize all variables throughout this paper using the characteristic length scale $\hat{l} = 50$~\textmu{}m, the viscosity of water $\mu = 0.001$~Pa\,s, and the torque scale $\hat{T}=6\times10^{-12}$~N\,m, which corresponds to the magnitude of magnetic torques applied in similar experiments. \cite{nondimensional1} To describe the fluid flow around and driven by a micro-robot and passive flexible filament, we consider a body force field $\vec{F}$ that is concentrated near the micro-robot and filament. This force field represents the stresses exerted on the fluid by the immersed objects. Specifically, we distribute $N$ nodes over the surface of the helical micro-robot and passive filament. Following the method of regularized Stokeslets,\cite{stokeslet, Stokeslet2} a regularized force is placed at each of these nodes. The total body force field evaluated at a point $\vec{x}$ is the sum of contributions of the form
\begin{equation}
\label{3}
  \vec{F_i}(\vec x) = \vec{f_i} \phi_\epsilon (\vec{x}-\vec{x}_i),  
\end{equation}
where $\vec{f}_i$ is the force applied around the $i$th node, with position $\vec{X}_i$.
% \begin{align}
%   \vec{F}(\vec x) = \vec{f}(\vec y) \phi_\epsilon (\vec{x}-\\ \vec{y})\\
%    \vec{T}(\vec x) = \vec{t}(\vec y) \phi_\epsilon (\vec{x}-\\ \vec{y}).
%\end{align}
The regularization function is chosen to be\cite{Stokeslet2}
\begin{equation}
\label{4}
\phi_\epsilon(\vec r) = \frac{15\epsilon^4}{8\pi(r^2+\epsilon^2)^{7/2}},
\end{equation}
where $r=||\vec{r}||$, and $\epsilon$ is the regularization parameter, which controls the distance over which the force is concentrated. The velocity at a point $\vec{x}$ due to $N$ regularized forces $\vec{f_i}$ applied at positions $\vec{x_i}$ is\cite{stokeslet}
\begin{equation}
\label{5}
\vec{u}(\vec{x}) = \frac{1}{8\pi}\sum_{i=1}^N \frac{(r^2+2\epsilon^2)\vec{f_i}+[\vec{f_i}\cdot(\vec{x}-\vec{x_i})](\vec{x}-\vec{x_i})}{(r^2+\epsilon^2)^{3/2}},
\end{equation}
%and the velocity due to a regularized torque ${\vec{T}}_i$ at $\vec{X}_i$, is given by
%\begin{equation}
%\vec{u}(\vec{X}) = \frac{1}{16\pi}\frac{(2r_i^2+5\epsilon^2)({\vec{T}}_i\times(\vec{X}-%\vec{X}_i))}{(r_i^2+\epsilon^2)^{5/2}},
%\end{equation}
where $r_i=||\vec{x}-\vec{x}_i||$.

The centerline equation of the helical micro-robot in the initial configuration is defined as:
\begin{align}
\label{9}
x(s) &= R \cos{s},\notag\ \\
y(s) &= \pm R \sin{s},\notag\ \\
z(s) &= bs,
\end{align}
where $0\leq s \leq s_{\max}$, and the radius and pitch of the helix are $R$ and $\lambda = 2\pi b$, respectively. The positive sign in the equation for $y$ is used for right-handed helices and the negative sign is used for left-handed helices. The arc length of the helical micro-robot is equal to $s_{\max}\sqrt{R^2+b^2}$. For the filament, we consider a straight initial configuration, coinciding with the rest configuration.

Mesh nodes are distributed over the filament and the micro-robot by discretizing the objects into many cross-sections along their respective centerlines and further dividing the circumference of each cross-section into multiple nodes (three for the helix and four for the filament). We choose the number of cross-sections of the helix and filament depending on their lengths and cross-sectional radii to ensure that the spacing between neighboring cross-sections and the spacing between nodes at a cross-section are equal. The cross-sections are placed along and perpendicular to the centerline of the filament and the axis of the helical micro-robot. Throughout this study, we set independent values for the regularization parameter on the filament and the micro-robot, equal to the minimum distance between nodes on their respective surfaces.

To model the flexible filament, each filament node is connected to the other nodes at the same cross-section and at neighboring cross-sections by Hookean springs with a spring constant of $k$, which provide the desired rigidity to the overall structure. At each time step, the elastic forces due to these springs are calculated based on the instantaneous configuration of the nodes.

In addition to the hydrodynamic forces, a repulsive steric force is introduced to prevent collisions between the filament and the micro-robot. When the nodes of the filament and the micro-robot are in close proximity, a truncated Lennard-Jones potential is employed to exert this repulsive force \cite{repulsiveForce}:
\begin{equation}
\label{6}
E_{LJ}(\vec{r_s}) = \frac{\sigma F_s}{6}\left[\left(\frac{\sigma}{|\vec{r_s}|}\right)^{12}-\left(\frac{\sigma} {|\vec{r_s}|}\right)^6\right]H(2^\frac{1}{6}\sigma-\vec{r_s}).
\end{equation}
where $H$ is the Heaviside step function, $\vec{r_s}$ is the vector connecting the two nearby points, $\sigma$ is the cutoff distance, and $F_s$ sets the repulsion strength. Then, the repulsive force can be computed by
\begin{equation}
\label{7}
\left\{
\begin{alignedat}{3}
% R & L   &  R & L   &  R & L 
\vec{F}^{steric} = -\nabla E_{LJ}, \qquad{} |\vec{r_s}|<\sigma, \\
\vec{F}^{steric} = \vec{0}, \qquad{} |\vec{r_s}|\geq\sigma.
\end{alignedat}
\right.
\end{equation}
We compute the hydrodynamic force $\vec{f}_i$ at filament nodes by applying a force balance condition, namely, that the hydrodynamic force balances the steric and elastic forces.

The helical micro-robot, in contrast to the filament, is simulated as a rigid body; the forces applied at the helix nodes are determined so that the nodal velocities, computed via \eqref{5}, are consistent with a rigid body translation and rotation. Additionally, the total hydrodynamic force exerted by the helix balances the steric repulsion force and the total hydrodynamic torque balances the magnetic torque $\vec{T}_M$ applied to the micro-robot, which is given by
\begin{equation}
\label{11}
\vec{T}_M = \vec{m}\times \vec{B},
\end{equation}
where $\vec{B}$ is the magnetic field and $\vec{m}$ is the magnetic dipole moment of the micro-robot, which we assume to be fixed in the (rotating) reference frame of the micro-robot. In the initial position, the magnetic moment of the micro-robot is aligned with the magnetic field in the $X$-direction. The rotating magnetic field is given by
\begin{equation}
\label{10}
    \vec{B}(t) = B_0 \begin{pmatrix}
    \cos{(2 \pi f t)} \\
    \pm \sin{(2 \pi f t)} \\
    0
\end{pmatrix},
\end{equation} 
where $f$ is the rotational frequency, $B_0$ denotes a constant magnitude of the magnetic field, and the direction of rotation is determined by the sign in the $Y$-component of \eqref{10}. The magnetic field rotates around the $\pm Z$-direction, which is parallel with the axis of the micro-robot in the initial configuration. 

In general, a magnetic torque acts to rotate a magnetic dipole to be aligned with the magnetic field. Since the magnetic field continuously rotates, the micro-robot rotates to follow the magnetic field, provided that the magnetic field rotates below a threshold known as the step-out frequency.\cite{HowShouldMicrorobotsSwim} For all simulations conducted in this paper, the rotational frequency presented in equation \ref{10} is $f=4$ and the  maximum possible magnitude of the non-dimensional magnetic torque is set to $B_0|\vec{m}|=50$.

Once all of the hydrodynamic forces on the micro-robot and filament nodes are determined, the velocities at the filament nodes are evaluated using equation \eqref{5}. These velocities, along with the rigid body translational and rotational velocity of the micro-robot, are used to update the node positions by the fourth-order Runge-Kutta method.

\section{Fluid transport due to prescribed rotation of the helical micro-robot}
\label{sec:prescribed_rotation}
In the first step, our objective is to gain a deeper understanding of the motion of the fluid near a rotating helical micro-robot. Two crucial ratios influence the flow field and movement of these robots: (1) $\lambda/R$ and (2) $r/R$. Here, $\lambda$, $R$, and $r$ denote the pitch, radius of the helical micro-robot, and the radius of its cross-section, respectively. We analyze the impact of the two ratios individually on pathlines in the fluid around the micro-robot. This knowledge will aid us in designing a helical micro-robot capable of efficiently transporting cargoes with flexible filaments. In the following section, we delve into studying the fluid's streamline.

To examine the effects of $\lambda/R$ and $r/R$, we maintain a constant helical micro-robot radius $R=0.1$ while varying $\lambda$ and $r$. Subsequently, we plot the pathline of the fluid. To mitigate any potential end effects caused by the helical micro-robot, we generate a long micro-robot with $N_\lambda = 10$ windings. A point is selected along the axis of the micro-robot, situated one quarter of the way down from the top of the helix. We apply a prescribed motion to the micro-robot so that it rotates around its axis without any translational velocity. The reason for considering this prescribed motion rather than allowing free motion under the force-free and prescribed torque conditions is that we will later connect the helix to another helical section of opposite handedness, which will alter the translational velocity. 

We track and plot the pathline of the chosen point resulting from the rotation of the helix over time, using the velocity field given by \eqref{5}. The fluid pathline gives an indication of the motion of a point on a passive filament located along the axis of the micro-robot. Ensuring a strong coupling between the filament and the micro-robot is crucial for retaining the filament inside the micro-robot, as we will discuss in subsequent sections.

\begin{figure}[h!]
\centering
\includegraphics[width=\linewidth]{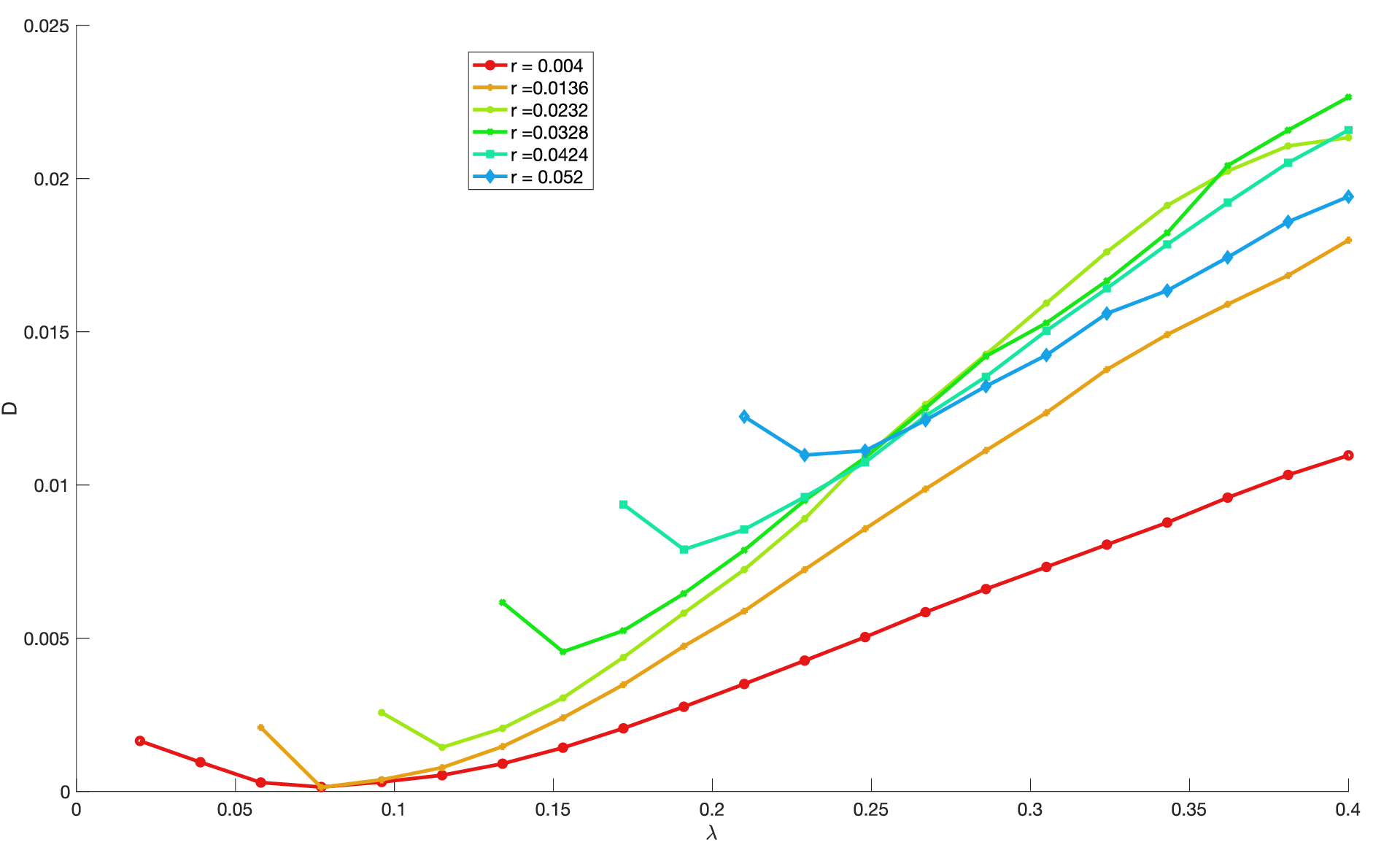}
\caption{The influence of $r$ and $\lambda$ on the maximum distance $D$ between the pathline of the point and the axis of the micro-robot. The radius of the helical micro-robot for all simulations is set to $R=0.1$.} 
\label{lambdaRrR}
\end{figure}

\begin{figure}[h!]
\centering
\includegraphics[width=\linewidth]{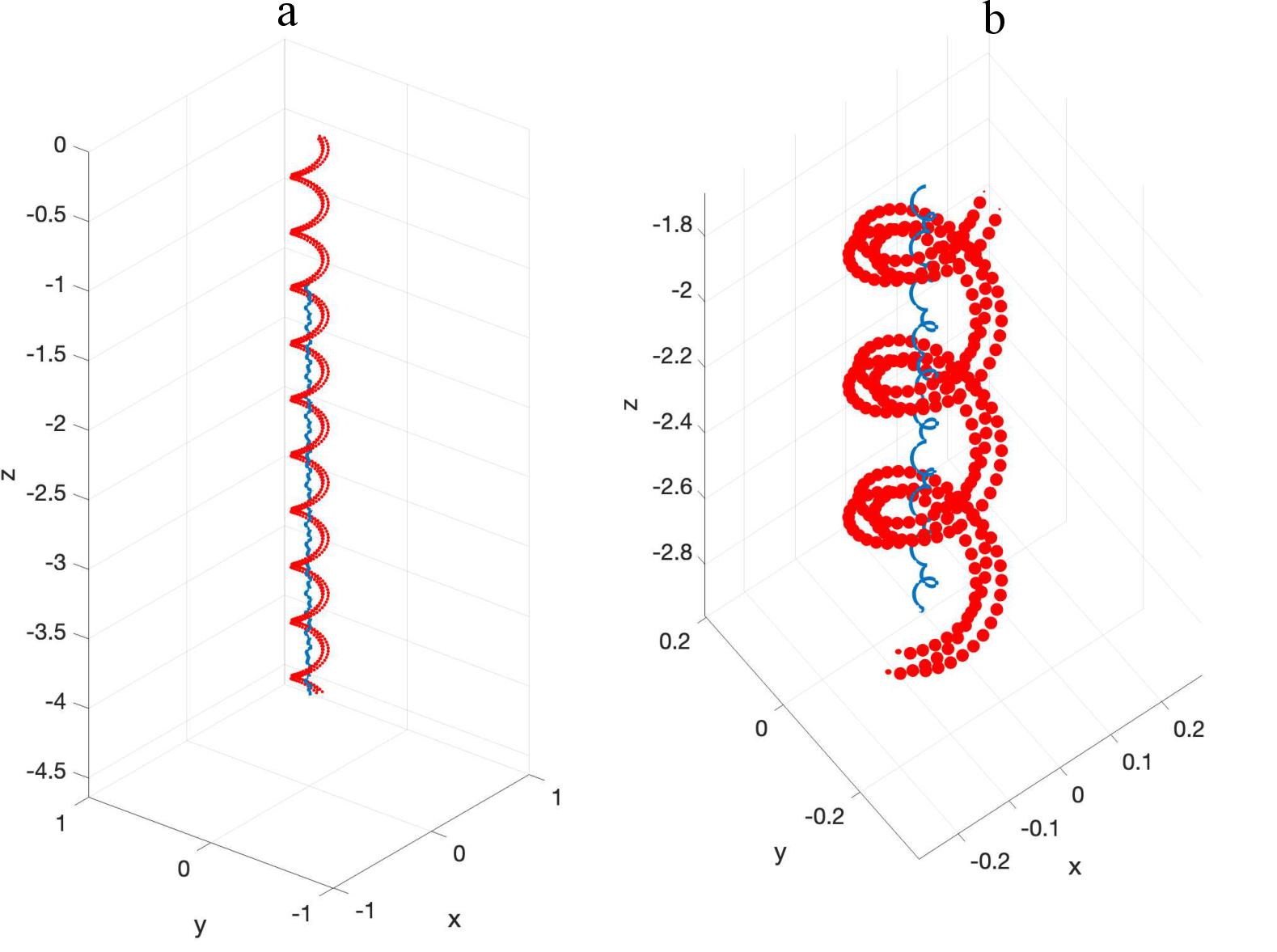}
\caption{Pathline of the point represented in 3D views with a close-up view of a portion in (b). The blue curve illustrates the trajectory of the point, while the red points depict the nodes situated on the surface of the micro-robot at one instant in time.} 
\label{pathline3D}
\end{figure}

Since our main focus is investigating how well a filament can be confined inside the micro-robot, we calculate the maximum distance ($D$) of the point from the axis of the micro-robot along its pathline as the point traverses from the top quarter plane to the bottom quarter plane (one quarter of the axial length of the helix from the bottom). The influence of $r$ and $\lambda$ on $D$ is shown in Figure \ref{lambdaRrR}. It can be observed that reducing the pitch of the micro-robot leads to a more confined pathline for the point, regardless of the value of $r$. However, no specific increasing or decreasing trend is observed for $D$ as $r$ increases. Nonetheless, it is evident that $D$ depends on both $r$ and $\lambda$. Figure \ref{pathline3D} showcases the 3D pathline of a point where the pitch of the micro-robot is $\lambda = 0.4$ and the radius of the cross section is $r=0.0232$. 


\section{The instantaneous flow field around a helical micro-robot in a rotating magnetic field}
\label{sec:helix_flowfield}
\begin{figure}[h!]
\centering
\includegraphics[scale=1]{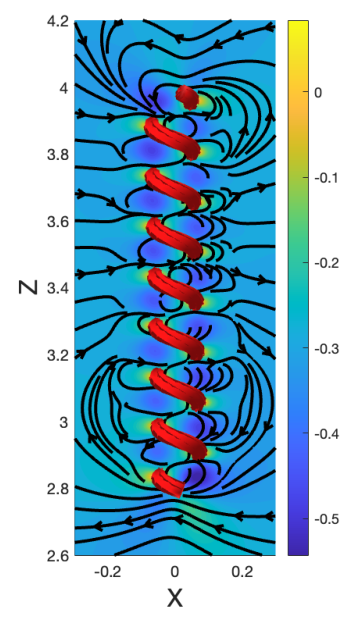}
\caption{Streamlines of the fluid in the $X$--$Z$ plane around a helical micro-robot in a rotating magnetic field. The color scale shows the velocity of the fluid in the $Z$ direction relative to the translational velocity of the micro-robot in the $Z$ direction ($V_z = 0.29$).} 
\label{VelSingleHelixStreamline}
\end{figure}

\begin{figure}[h!]
\centering
\includegraphics[width=\linewidth]{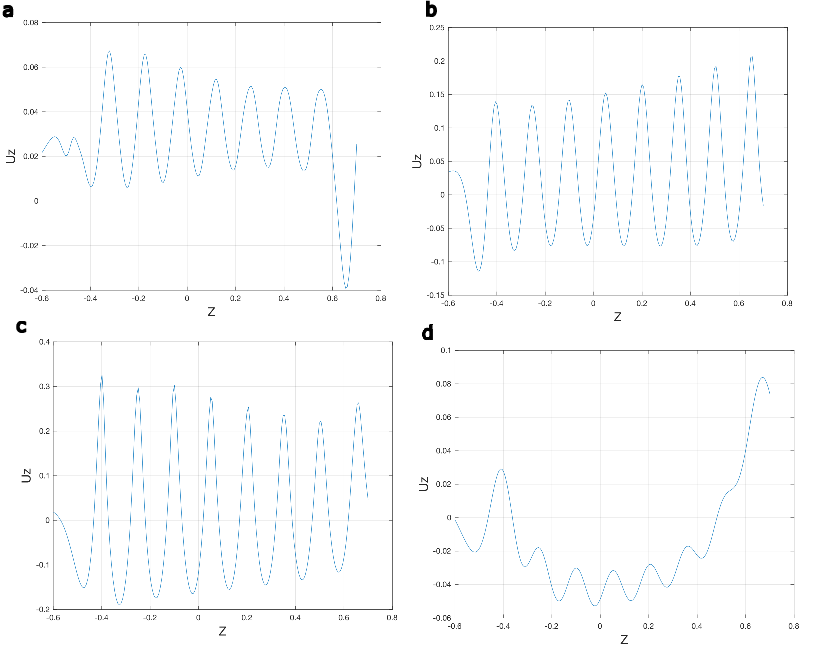}
\caption{Velocity profile of the fluid in the $Z$ direction versus $Z$ position in the $X$--$Z$ plane at a) $X=0$ (on the axis) b) $X=0.02$ (between the axis and the surface of the micro-robot) c) $X=0.0375$ (close to the surface of the micro-robot) and d) $X=0.1$ (beyond the surface of the micro-robot). The helix parameters are $\lambda = 0.15$, $R=0.06$, and $r=0.015$. The top and bottom of the micro-robot are located at $Z=0.6$ and $Z=-0.6$, respectively. } 
\label{velSingleProfileRev}
\end{figure}

To complement the pathlines for prescribed motion of the micro-robot, we now consider a force-free helical micro-robot driven by a rotating magnetic field and plot the fluid streamlines in the $X$--$Z$ plane (Figure \ref{VelSingleHelixStreamline}). We use $N_\lambda=8$ windings, pitch $\lambda = 0.15$, radius $R=0.06$, and cross-sectional radius $r = 0.015$ for the helix. The instantaneous streamlines have a complex pattern with significant lateral motion, which is expected to cancel out on average over a revolution of the micro-robot. The vertical velocity is almost everywhere negative (in the backward direction) relative to the micro-robot, and most negative between the turns of the microhelix at distances $\approx R$ from the helical axis. Therefore, in order to improve the coupling between the cargo and the micro-robot, it is crucial to keep the cargo inside the micro-robot. The ability to keep the pathline of a point inside the micro-robot, as discussed in the previous section, depends on the geometry of the micro-robot. Additionally, it is demonstrated in the following section that the flexibility of the filament is another parameter that affects the behavior of the passive filament inside the micro-robot.
To better understand the flow around the micro-robot, we also plot the $Z$-component of the velocity field as a function of $Z$ along four lines at different distances from the $Z$ axis in the $X$--$Z$ plane (Figure \ref{velSingleProfileRev}). The $Z$-velocity of the fluid on the $Z$-axis ($X=0$) is mostly positive, i.e., in the forward direction, but with a low speed $U_z \sim 0.04$ compared with the non-dimensional translational speed of the micro-robot $V_z = 0.29$. However, as one moves away from the axis, the $Z$-velocity of the fluid oscillates along $Z$ with a larger amplitude and alternates between forward and backward directions. 

Furthermore, Figure \ref{velSingleProfileRev}d clearly illustrates that as one moves beyond the helical micro-robot, not only does the velocity of the fluid in the $Z$-direction reduce, but a larger region experiences velocities in the backward direction. To prevent decoupling between the filament and the micro-robot, it is desirable for the filament to remain in a location where the fluid is moving forward at the same speed as the helix. Therefore, it is crucial to carefully design the micro-robot to ensure that the filament stays inside the micro-robot and has maximum coupling with the micro-robot.

\section{Coupling between a passive filament and a helical micro-robot}
\label{sec:one_helix}
In this section, we examine the motion of a passive flexible filament, of length $L_f=1$ and radius $r_f = 0.014 $, located along the axis of a rotating magnetic helical micro-robot. Our objective is to investigate the coupling between the micro-robot and the passive filament in such an arrangement. In the preceding sections, we studied the streamlines and the pathlines of the fluid surrounding the helical micro-robot. Based on the obtained streamlines, it is critical to confine the filament inside the micro-robot to apply hydrodynamic drag force in the forward direction and establish good coupling between the filament and the micro-robot. Moreover, based on the pathline studied in the previous section, selecting the $r/R$ and $\lambda/R$ ratios carefully is crucial to confine the pathline of fluid inside the micro-robot. We consider the baseline parameter values of helical radius $R=0.06$, axial length of the micro-robot $L=1.2$, cross-sectional radius of the micro-robot $r=0.015$, helical pitch $\lambda = 0.2 $ (corresponding to six windings), and spring constant between filament mesh nodes $k=1000$. From these values, we separately vary $\lambda$, $k$, and $r$. To characterize how well the micro-robot and filament move together, we initialize the micro-robot and filament so that their top ends are at the same $Z$ position. We then track their relative displacement, 
\begin{equation}
\label{5.1}
%\chi =\frac{Z_{helix}-Z_{filament}}{Z_{helix}},
\Delta Z = Z_{helix}-Z_{filament},
\end{equation}
until a fixed dimensionless time $t_{end}=4.8$. Although it is possible to transport the filament with any displacement $\Delta Z$ as long as it becomes steady at long times, we aim for $\Delta Z = 0$ so that the filament is at the top of the micro-robot (which may help with releasing the cargo at the destination) without protruding from the microhelix. We anticipate that keeping the filament confined within the micro-robot will allow the micro-robot to be turned to a different swimming direction without losing the filament.

\subsection{The impact of helical pitch of the micro-robot}
\label{subsec:pitch}
\begin{figure}[h!]
\centering
\includegraphics[width=\linewidth]{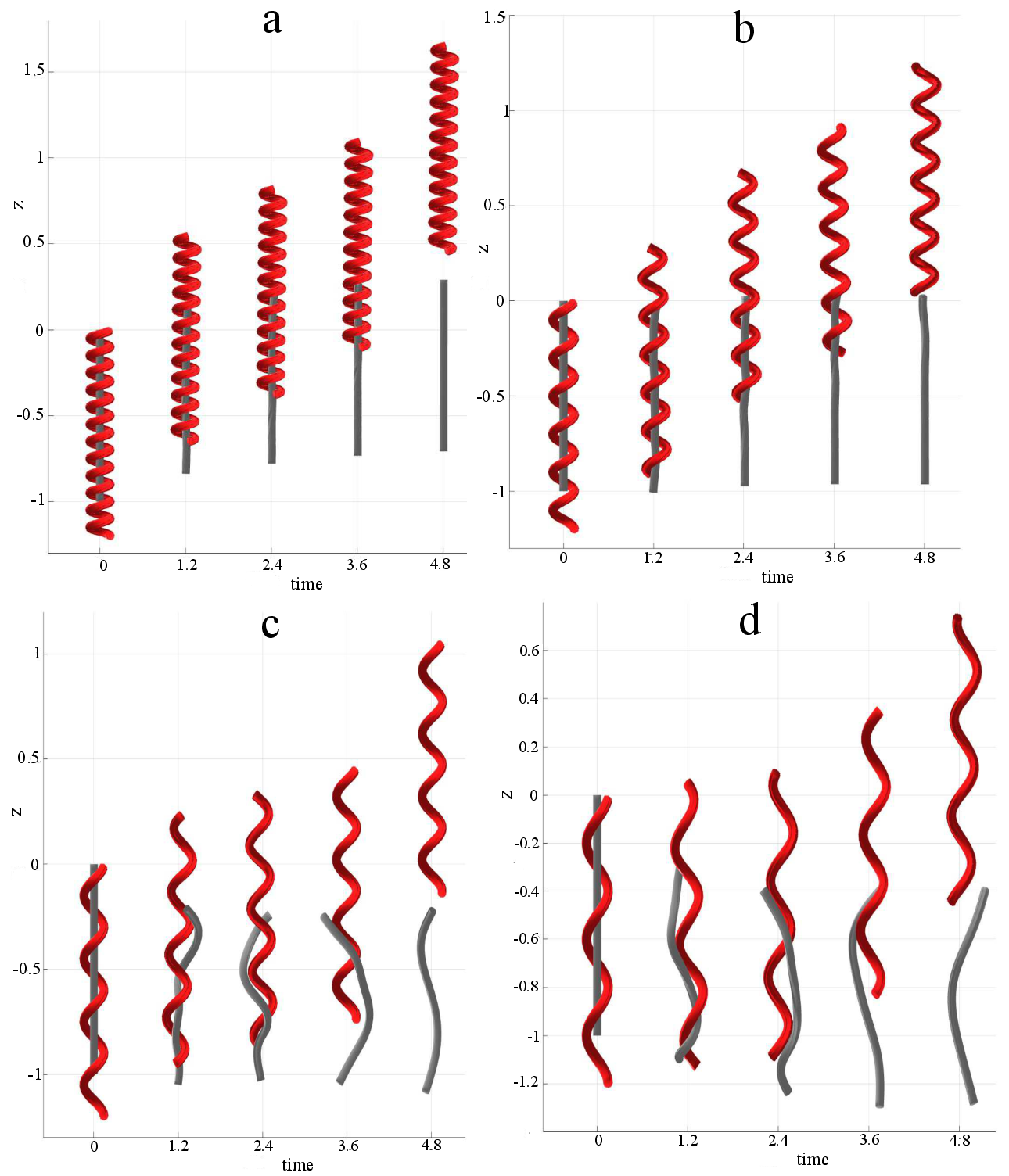}
\caption{Snapshots of the motion of a helical micro-robot and a passive filament with different ratios of helical pitch to radius ($\lambda/R$) a) 1.6 b) 3.3 c) 5 d) 6.6.} 
\label{LambdaPos}
\end{figure}

The motion of the micro-robots and filaments are presented visually in Figure \ref{LambdaPos} for four values of $\lambda/R$. In all cases, the micro-robot moves faster than the filament so the two objects decouple. Using the relative displacement $\Delta Z$ as an indication of coupling, we found that performance was similarly poor with all helical pitches ($\Delta Z \approx L$ at time $t=4.8$) and there was no clear trend with $\lambda/R$. In terms of the absolute motion, for small $\lambda/R$, the filament moves in the forward direction whereas for large pitches ($\lambda/R = 3.3$ and greater), the filament has a net downward displacement. For larger $\lambda/R$, the helix is not able to confine the filament, which bends significantly and moves laterally to the exterior of the micro-robot. We note that the $Z$-velocities of the fluid at these larger distances from the helical axis are more negative (Figure~\ref{velSingleProfileRev}), which explains why the filament moves backward in these cases. Therefore, careful selection of the geometry of the helix is necessary to achieve good coupling.  
%We found the best coupling ($\Delta Z$ closest to 0 at time $t=4.8$) between the helix and filament occurs for $\lambda/R = 6.6$.

\subsection{The impact of flexibility of the passive filament} 
\begin{figure}[h!]
\centering
\includegraphics[width=\linewidth]{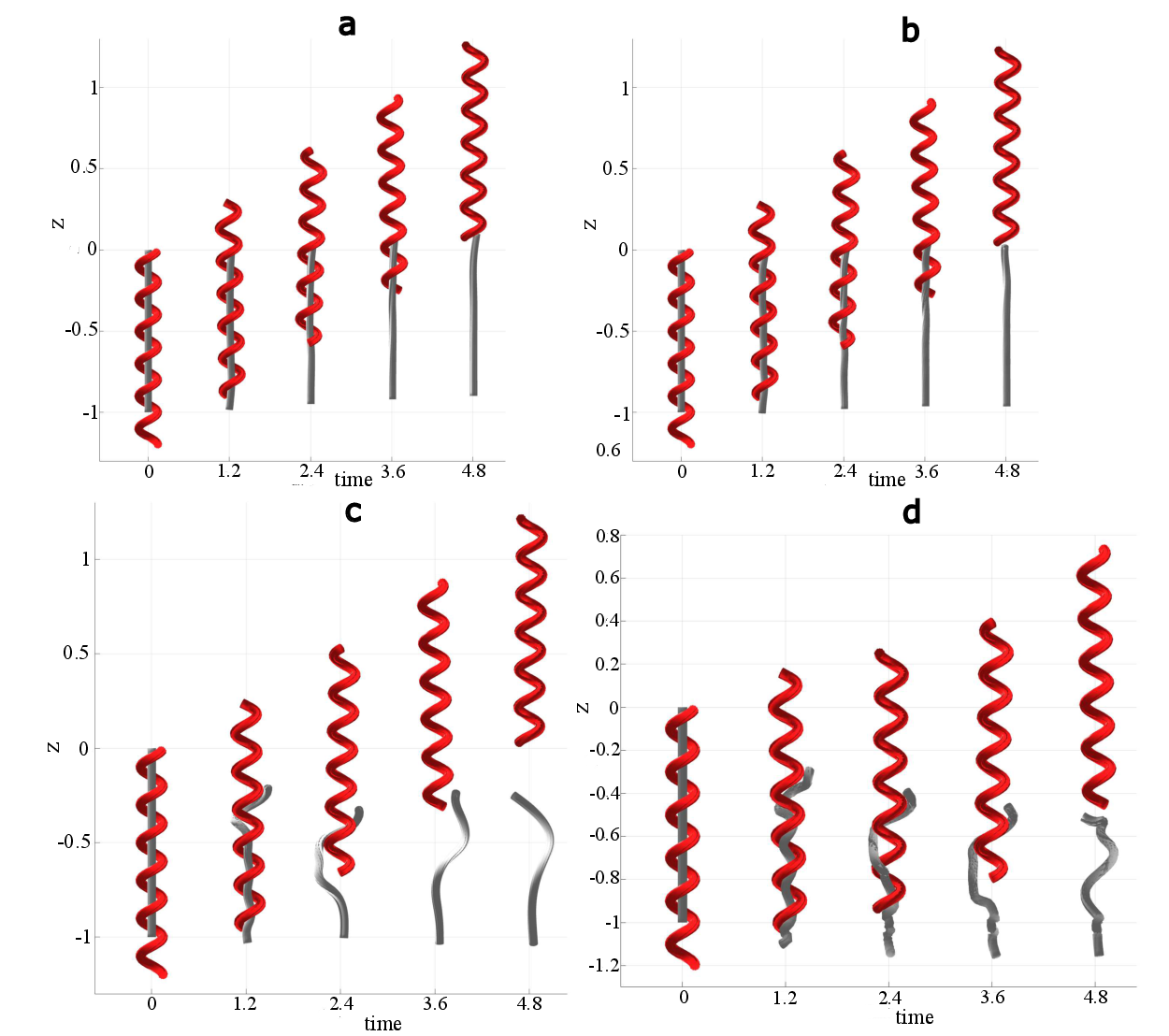}
\caption{Snapshots of the motion of the helical micro-robot and the passive filament with different flexibility (Hookean spring constant between mesh nodes of the filament) a) $k=1000$ b) $k=300$ c) $k=50$ d) $k=5$. } 
\label{FlexPos}
\end{figure}

We now vary the flexibility of the filament to study its effect on the coupling between the filament and the micro-robot. As shown in Figure \ref{FlexPos}, when the filament is very flexible, it undergoes high deformation, causing some parts to extend beyond the surface of the micro-robot. As previously discussed, this results in the filament being pushed backwards. On the other hand, stiff filaments are confined closer to the axis where they are driven forward by the flow field. Hence, coupling is best for stiffer filaments. 

\subsection{The impact of cross-sectional radius of the micro-robot} 
In this section, the ratio of $r/R$ is varied, for constant helical radius $R=0.06$, to investigate its impact on the motion of the helical micro-robot and the filament. Four different values for $r/R$ are tested: 0.25, 0.2, 0.16, and 0.13. The obtained results indicate that when $r/R$ is set to 0.25, the value of $\Delta Z$ is approximately 1.05. This means that at time 4.8, the top of the micro-robot is positioned approximately 1.05 units (almost the full axial length of the micro-robot) ahead of the top of the filament. For $r/R$ values of 0.2, 0.16, and 0.13, the corresponding $\Delta Z$ values are 1.15, 1.24, and 1.25, respectively. We remark that increasing $r/R$ reduces the space available within the helix, forcing the filament to be in closer proximity to the micro-robot. Viscous effects then drag the filament with a velocity closer to that of the micro-robot, leading to a better coupling between the filament and the micro-robot. However, $\Delta Z \approx L$ in all cases, indicating that the micro-robot eventually overtakes the filament, and the coupling is lost. In the next section, we will attempt to redesign the micro-robot to achieve complete coupling between the filament and the micro-robot.

\section{Improved design with multiple helical sections }
\subsection{Micro-robot with two helical sections}
\label{subsec:two_helix}
\begin{figure}[h!]
\centering
\includegraphics[scale=0.4]{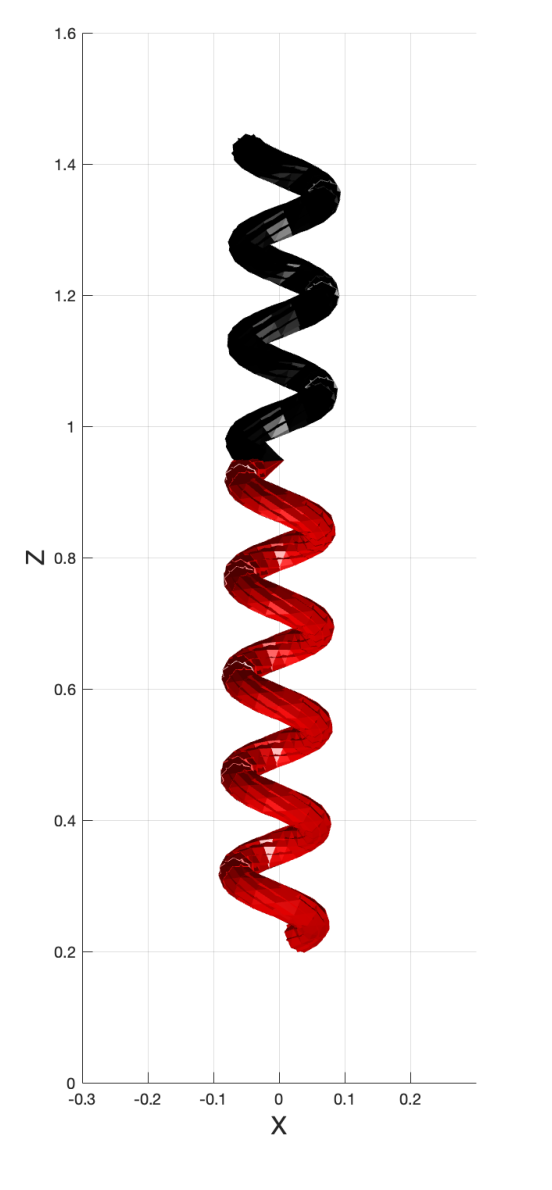}
\includegraphics[scale=0.8]{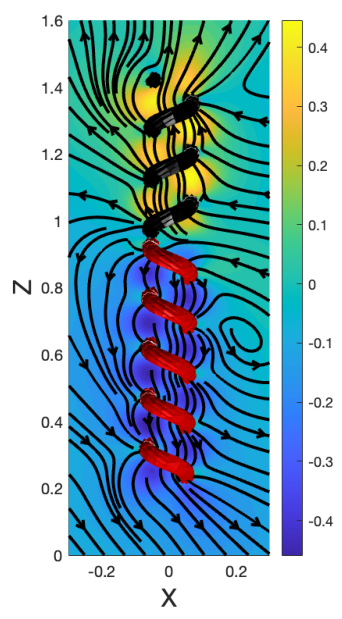}
\caption{A micro-robot consisting of a pushing (black) and pulling (red) microhelix. Fluid streamlines around the rotating micro-robot are shown on the right, with the color scale indicating the $Z$-component of the velocity relative to the $Z$-velocity of the micro-robot.} 
\label{TwoHelix}
\end{figure}

As demonstrated in the preceding sections, the simple helical micro-robot is able to propel the filament forward if the geometrical parameters are suitably chosen. However, the average speed of the filament is always lower than that of the micro-robot, hence, the micro-robot eventually overtakes the filament and leaves it behind. The objective of this project is to devise a novel design that enhances the coupling between the micro-robot and the filament. To achieve this, we must decrease the velocity difference between the micro-robot and the filament. As an initial approach, two microhelices, one pushing and one pulling, as depicted in Figure \ref{TwoHelix}, are connected in series. Model parameters are as in Section~\ref{sec:one_helix}.
The presence of a pushing helix at the front of the micro-robot reduces the speed of the micro-robot while simultaneously increasing the speed of the filament by driving forward the section of the filament within the pushing section. 

Thus, we can improve the coupling between the helix and the micro-robot, while maintaining a positive swimming speed, by adjusting two parameters: the length ratio of the pulling section to the pushing section and the length ratio of the filament in the pushing section to that in the pulling section. 

\begin{figure}[h!]
\centering
\includegraphics[scale=0.5]{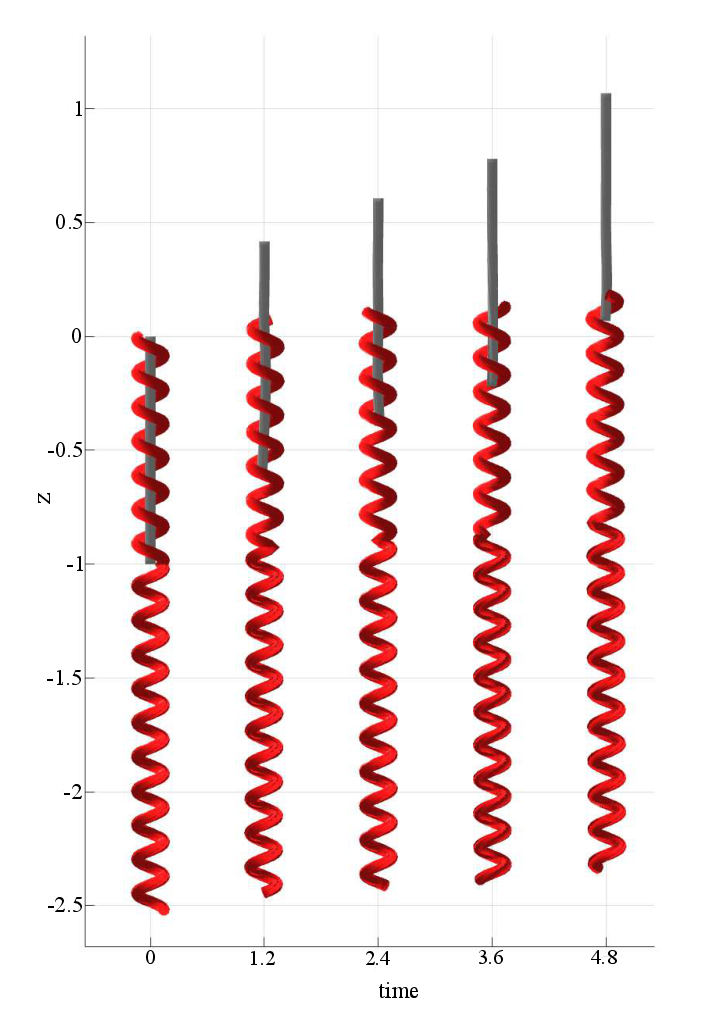}
\caption{A micro-robot with two microhelices initially containing a passive filament. The pushing section at the front of the micro-robot has the same axial length as the filament and pushes the filament out of the micro-robot.} 
\label{TwoHelixLong}
\end{figure}

The length ratio between the pushing and pulling sections of the microhelix can be adjusted to optimize the coupling between the micro-robot and filament. Figure \ref{TwoHelixLong} demonstrates that if the pushing microhelix is long enough to contain the filament, then the filament can move faster than the microhelix. Thus, by varying the length ratio between the pushing and pulling microhelix, along with the length of the filament within the pushing microhelix, the coupling can be significantly improved compared to a helical micro-robot with a single type of microhelix.

\begin{figure}[h!]
\centering
\includegraphics[scale=0.5]{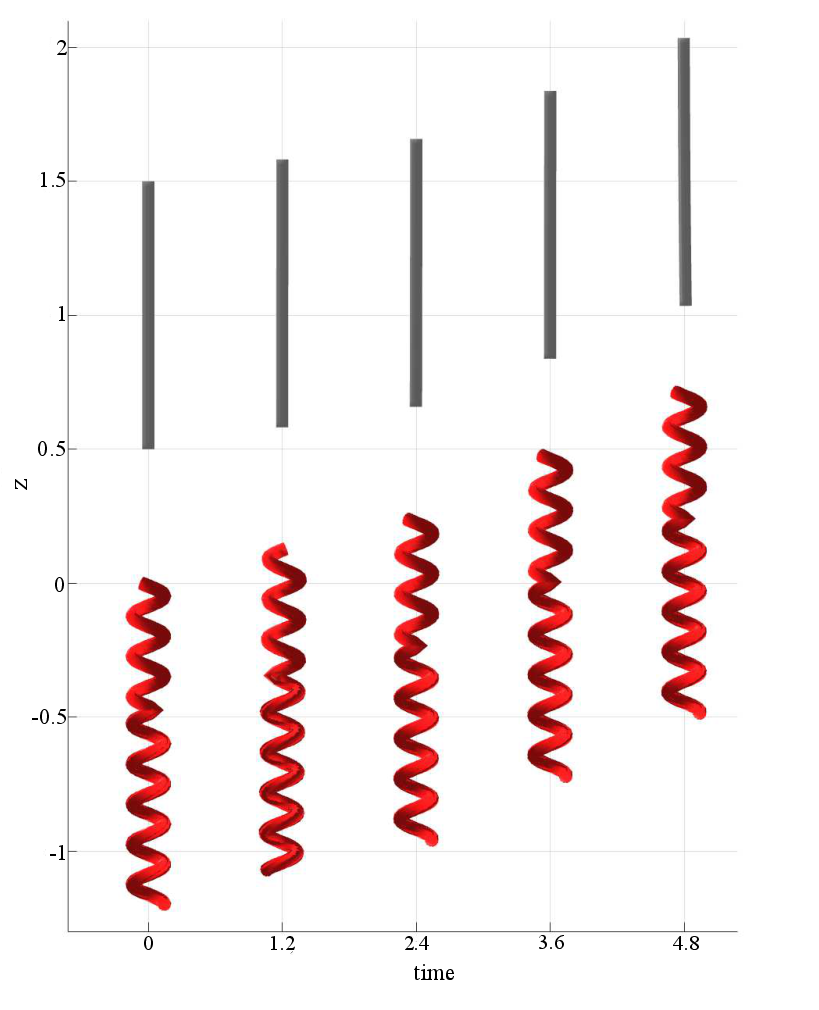}
\caption{A helical micro-robot with two microhelices approaching a passive filament.} 
\label{TwoHelixInteg}
\end{figure}

However, the initial step for carrying the filament is integrating (capturing) it with the micro-robot. Therefore, this study investigates the capability of the micro-robot to integrate with the filament. At the beginning, as shown in Figure \ref{TwoHelixInteg}, the filament is ahead of, and fully outside, the micro-robot. The micro-robot is manipulated using a magnetic field to approach the filament and pick it up. Different simulations are conducted to simulate the integration step. Figure \ref{TwoHelixInteg} shows that when the micro-robot approaches the filament, the filament is pushed forward, making it difficult for the micro-robot to catch it. This can be expected since the fluid moves forward in the $Z$-direction relative to the micro-robot, as shown in Figure \ref{TwoHelix}.

In contrast, the fluid velocity in front of the pulling helix is in a downward direction (Figure~\ref{VelSingleHelixStreamline}). This means that objects in front of the helix are drawn in, rather than being pushed away. To solve the integration issue in the current design, a pulling helix is added to the front of the micro-robot. Thus, the final design, which is studied in the next section, is constructed from three helices, including two pulling and one pushing helix. The pushing helix is located in the middle of the micro-robot between the two pulling helices with different lengths. Its length is carefully determined to optimize the coupling between the micro-robot and the filament.


\subsection{Micro-robot with three helical sections}
\label{sec:three_helix}
\begin{figure}[h!]
\centering
\includegraphics[scale=0.35]{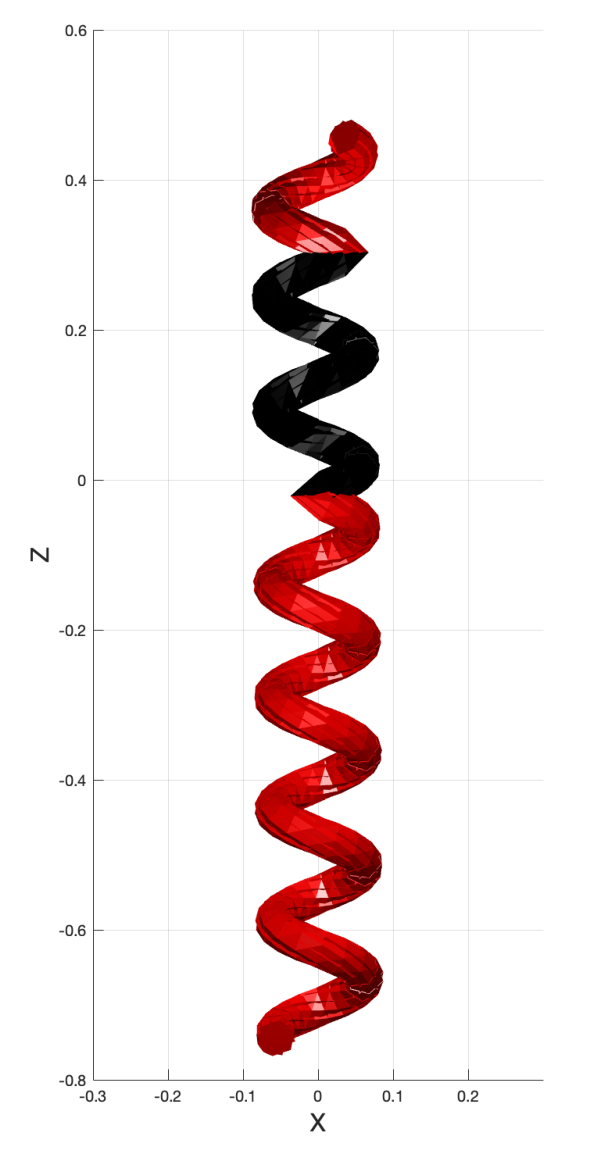}
\includegraphics[scale=0.7]{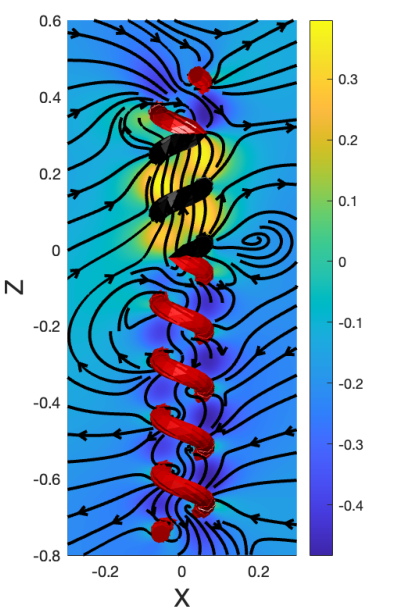}
\caption{A micro-robot consisting of three microhelices: a pushing (black) section between two pulling (red) sections. Fluid streamlines around the rotating micro-robot are shown on the right, with the color scale indicating the $Z$-component of the velocity relative to the $Z$-velocity of the micro-robot.} 
\label{ThreeHelix}
\end{figure}

To improve the integration process of micro-robot and the filament, a pulling helix is added to the front of the micro-robot discussed in Section~\ref{subsec:two_helix}. A schematic of the new micro-robot, and the generated flow field, is presented in Figure \ref{ThreeHelix}. 
The middle section of the micro-robot generates the necessary drag force to propel the filament forward, while the front section draws the filament into the helix to aid the integration process. The back part of the micro-robot generates most of the driving force required for the micro-robot to move forward. The length ratio between the middle, front, and back sections of the micro-robot must be carefully chosen to maximize the coupling between the filament and the micro-robot while providing forward propulsion for the coupled structures.

In Section~\ref{sec:analytical}, we use a simplified analytical model to estimate the swimming speeds and show that full coupling can be achieved for many combinations of helical pitch and cross-sectional radius. It is important to note that various design criteria must be considered when designing the micro-robot. Firstly, increasing the total length of the micro-robot may enhance the speed of the filament and the micro-robot, but also reduces the robot's manoeuvrability as it will be unable to pass through narrow channels with sharp bends. Thus, the total length of the micro-robot must be specified based on its intended application. The second criterion to consider is the position of the end part of the filament, which should be located inside the middle (pushing) part of the micro-robot for stability of the coupled configuration. These two considerations are also discussed in Section~\ref{sec:analytical}.

We fix the total length of the micro-robot $L=2.5$ and let the sum of the lengths of the front and middle microhelices be $1.125$, which is slightly greater than the length of the filament $L_f=1$. The ratio of the middle (pushing) length to the combined front and middle length, denoted $\xi$, is varied from 0.3 to 0.6.

\begin{figure}[h!]
\centering
\includegraphics[width=\linewidth]{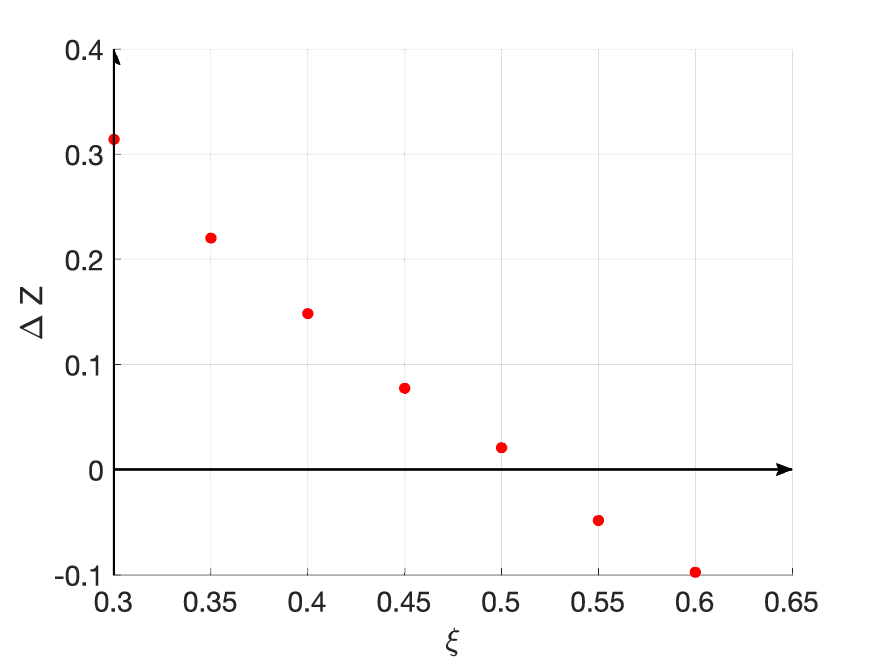}
\caption{The effect of length ratio $\xi$ of the middle (pushing) length to the combined front and middle lengths on the coupling between the three-section helical micro-robot and the passive filament.} 
\label{ThreeHelixRatioLength}
\end{figure}

\begin{figure}[h!]
\centering
\includegraphics[scale=0.45]{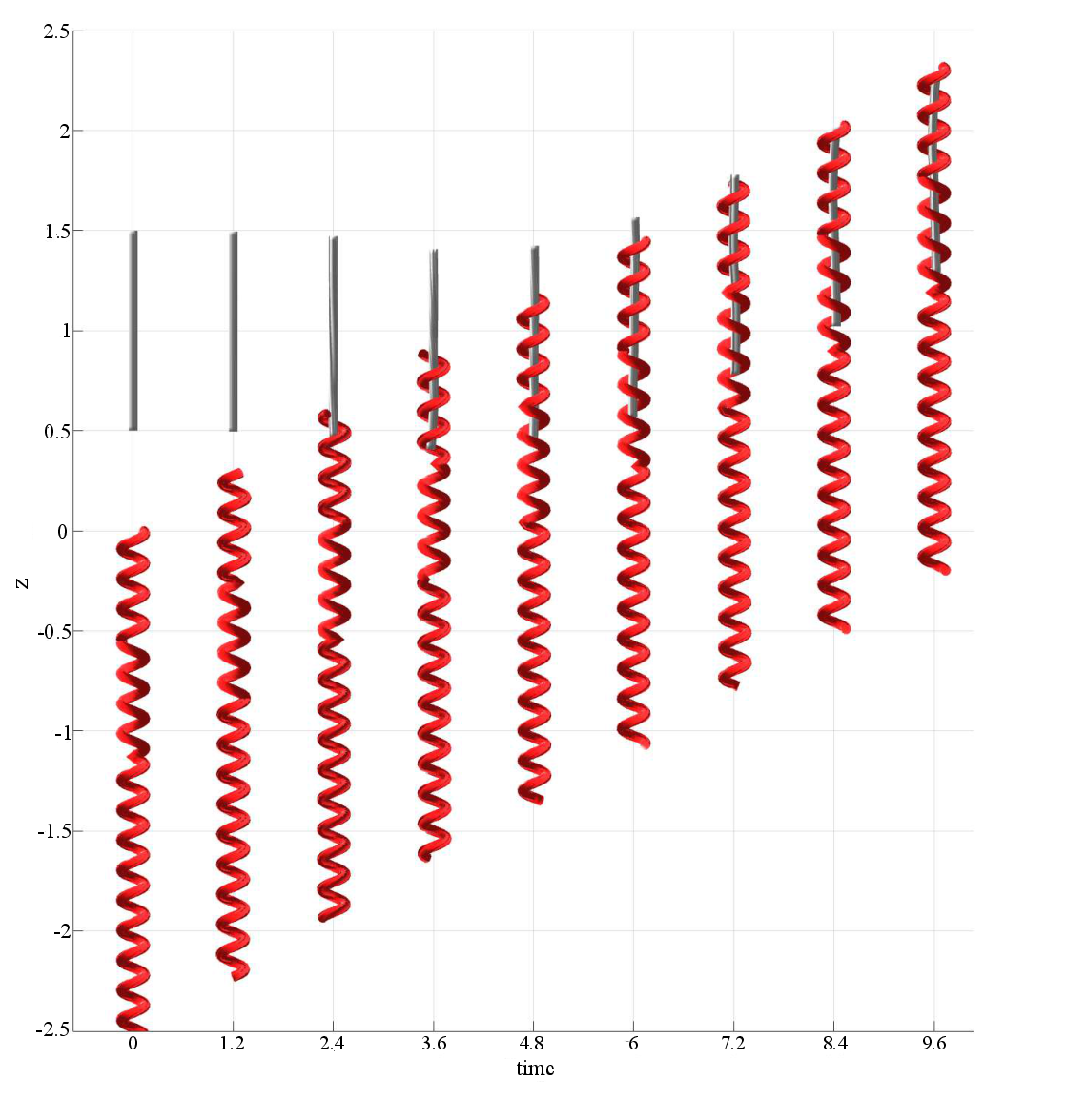}
\caption{Integration and carrying the passive filament with the designed helical micro-robot with three microhelices.} 
\label{FullyCoupling}
\end{figure}

As seen in Figure \ref{ThreeHelixRatioLength}, when the pushing length ratio is $\xi = 0.3$, the filament moves slower than the robot because the pushing microhelix is much shorter than the pulling microhelices. Consequently, full coupling between the micro-robot and the filament is not achieved. Increasing $\xi$ increases the speed of the filament. By interpolation of the approximately linear trend shown in Figure \ref{ThreeHelixRatioLength}, the best length ratio $\xi$ for fully coupling ($\Delta Z = 0$) was determined to be approximately 0.52. The processes of integration and coupling for the microrobot with the length ratio $\xi=0.52$ are simulated by initializing the filament ahead of the micro-robot. As shown in Figure \ref{FullyCoupling}, both of these steps are successful. In summary, the design's advantage lies in the pulling microhelix as the front microhelix, which causes the filament to be pulled toward the micro-robot during the integration phase and prevents the filament from moving ahead of the micro-robot, facilitating coupling. Furthermore, by choosing the lengths of the front and middle sections so that part of the pushing microhelix is below the bottom of the filament, we prevent the filament from falling behind. Therefore, not only can coupling and integration be achieved with the current design, but the coupled configuration is stable.

\section{Analytical study on the possibility of coupling }
\label{sec:analytical}
We develop a reduced analytical model to demonstrate, with a small number of parameters that must be determined numerically, that full coupling between the micro-robot and passive filament is generically achievable by changing the proportions of pushing and pulling helices. Consider a long right-handed microhelix. Adopting a resistive force theory approach (in particular, ignoring end effects), the relationships between the axial velocity $U$ of the helix, the rotational rate $\Omega$ of the helix, the thrust force per unit length $F$ applied to the fluid, and torque per unit length $T$ (in the forward axial direction) applied to the fluid, can be expressed as\cite{helixMotionFormula, helixMotionFormula2mf}

\begin{equation}
\label{7.1}
\left\{
\begin{alignedat}{3}
% R & L   &  R & L   &  R & L 
 A_{11}{}U +{} A_{12}{}\Omega = F, \\
 A_{21}{}U +{} A_{22}{}\Omega = T,
\end{alignedat}
\right.
\end{equation}
where the resistance matrix per unit length  $\begin{pmatrix}
    A_{11} & A_{12} \\
    A_{21} & A_{22}
\end{pmatrix}$ is a symmetric matrix that depends only on the cross-sectional radius, helical radius, and pitch of the microhelix. A left-handed microhelix with the same geometric parameters would have the same resistance coefficients except for a change in signs for the off-diagonal elements.\cite{helixMotionFormula} For motion driven by an external torque $T$, we set $F=0$ and express the translational velocity of the helix as
\begin{equation}
    U = -\frac{A_{12}}{A_{11}}\Omega.
\end{equation}
It should be noted that $A_{12}/A_{11}$ is negative for a right-handed helix so that $U>0$ (pulling) if $\Omega > 0$. In our proposed geometry, the micro-robot comprises both a pulling microhelix and a pushing microhelix, which are of opposite handedness. Denoting the ratio of the pushing microhelix length to the total length by $\eta$, the forward swimming speed of the compound micro-robot is
\begin{equation}
    U = -\frac{A_{12}}{A_{11}}\Omega(1-2\eta).
    \label{eq:robot_velocity}
\end{equation}

\begin{figure}[h!]
\centering
\includegraphics[scale=0.6]{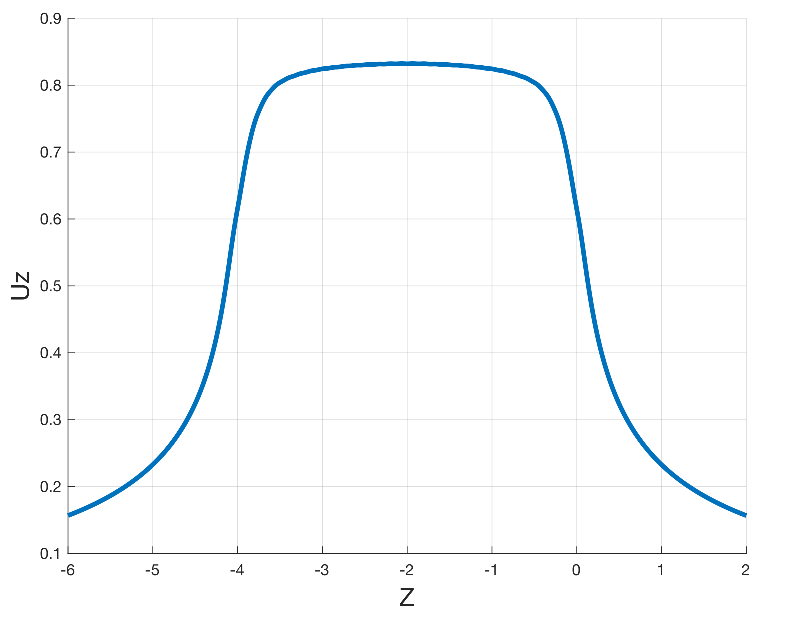}
\caption{Velocity profile of the fluid along the axis of a right-handed helix in the axial direction when the microhelix moves with prescribed speed $U=1$ and $\Omega = 0$. The back and front of the microrobot are located at $Z=-4$ and $Z=0$, respectively. The velocity at the central plateau corresponds to the parameter $\alpha$.} 
\label{U1}
\end{figure}

\begin{figure}[h!]
\centering
\includegraphics[scale=0.6]{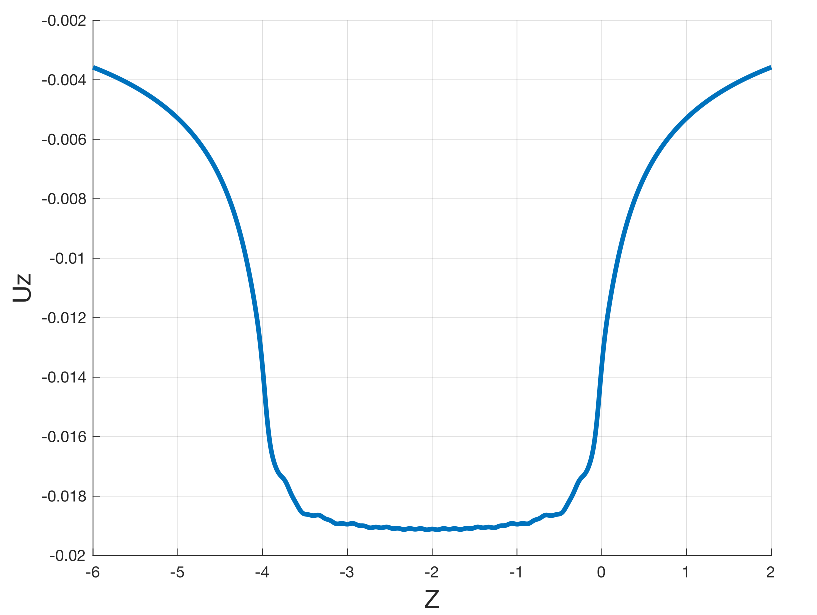}
\caption{Velocity profile of the fluid along the axis of a right-handed helix in the axial direction when the microhelix moves with prescribed speed $U=0$ and $\Omega = 1$. The back and front of the micro-robot are located at $Z=-4$ and $Z=0$, respectively. The velocity at the central plateau corresponds to the parameter $\beta$.} 
\label{Omega1}
\end{figure}

To estimate the velocity of the passive filament, we formulate equations analogous to \ref{7.1} for the passive filament, where the force and torque densities are interpreted as the hydrodynamic force and torque densities that would act on a stationary filament in the flow field generated by the helical micro-robot. In general, these quantities depend on all of the geometric parameters of the problem as well as the position of the filament relative to the helix. For long helices, we find that the fluid velocity in the axial direction is roughly uniform along the axis of a rotating or translating helix, away from the ends (Figures ~\ref{U1} and \ref{Omega1}). Hence, we assume that the filament lies along the helical axis and is nested within the axial bounds of the micro-robot. We then make the approximation that the force per unit length and the torque per unit length on the filament are proportional to the uniform fluid velocity along the axis of the helix, which is in turn linearly related to the rotational rate and translational velocity of the helix. The system of equations governing the passive filament inside the right-handed helix can be expressed as
\begin{equation}
\label{7.2}
\left\{
\begin{alignedat}{3}
% R & L   &  R & L   &  R & L 
 B_{11}{}U_f +{} B_{12}{}\Omega_f = (C_{11}U +{} C_{12}{}\Omega), \\
 B_{21}{}U_f +{} B_{22}{}\Omega_f = (C_{21}U +{} C_{22}\Omega).
\end{alignedat}
\right.
\end{equation}
Assuming that the filament is almost straight, the effect of $\Omega$ on the axial force on the passive filament is negligible, so $B_{12} \approx 0$. Therefore, the translational velocity of the filament is approximated by
\begin{equation}
\label{7.3}
    U_f = \alpha U + \beta \Omega,
\end{equation}
where $\alpha = C_{11}/B_{11}$ and $\beta = C_{12}/B_{11}$ are only functions of the cross-sectional radius, the helical pitch, and the helical radius of the micro-robot, which we determine numerically using the regularized Stokeslet method for a long helix (20 windings) with fixed helical radius $R=0.1$ and various combinations of cross-sectional radius $r$ and helical pitch $\lambda$. We expect that $\alpha$ takes values between 0 and 1, because the fluid should be dragged in the same direction as a translating helix but not faster than the helix. We expect $\beta < 0$ for right-handed helices, as demonstrated in Sections~\ref{sec:prescribed_rotation} and \ref{sec:helix_flowfield}.  For a left-handed helix, we would have the same values for $C_{11}$ and $B_{11}$, and hence $\alpha$. The coefficient $C_{12}$, and hence also $\beta$, would have the same magnitudes but opposite signs.

For a micro-robot with both left- and right-handed sections, the velocity of the filament is estimated by taking the average of the velocities of the portions of the filament in the right- and left-handed sections of the micro-robot, weighted by the lengths of these two portions. Suppose that the fraction of the filament contained in the left-handed section of the micro-robot is $\xi$ and the fraction in the right-handed section is $(1-\xi)$. In general, $\xi$ and $\eta$ may be different if the filament is shorter than the helix. If the filament runs through all of the left-handed portion of the micro-robot, then $\xi = \eta L/L_f$. For simplicity, we will assume that this is the case. The estimated velocity of the filament is
\begin{equation}
    U_f = \alpha U + \beta \Omega (1-2\eta L/L_f).
    \label{eq:filament_velocity}
\end{equation}

Coupling is achieved when the velocities of the micro-robot and the filament, respectively given by \eqref{eq:robot_velocity} and \eqref{eq:filament_velocity}, are equal. For nonzero rotation rate $\Omega$, there is a unique ratio 
\begin{equation}
    \eta^* = \frac{1}{2}\left[1-\frac{\left(\tfrac{L}{L_f}-1\right)\beta}{\tfrac{A_{12}}{A_{11}}(1-\alpha)+\beta\tfrac{L}{L_f}}\right],
    \label{eq:eta_star}
\end{equation}
at which $U=U_f$ for given helical parameters.  

\begin{figure}[h!]
\centering
\includegraphics[scale=0.3]{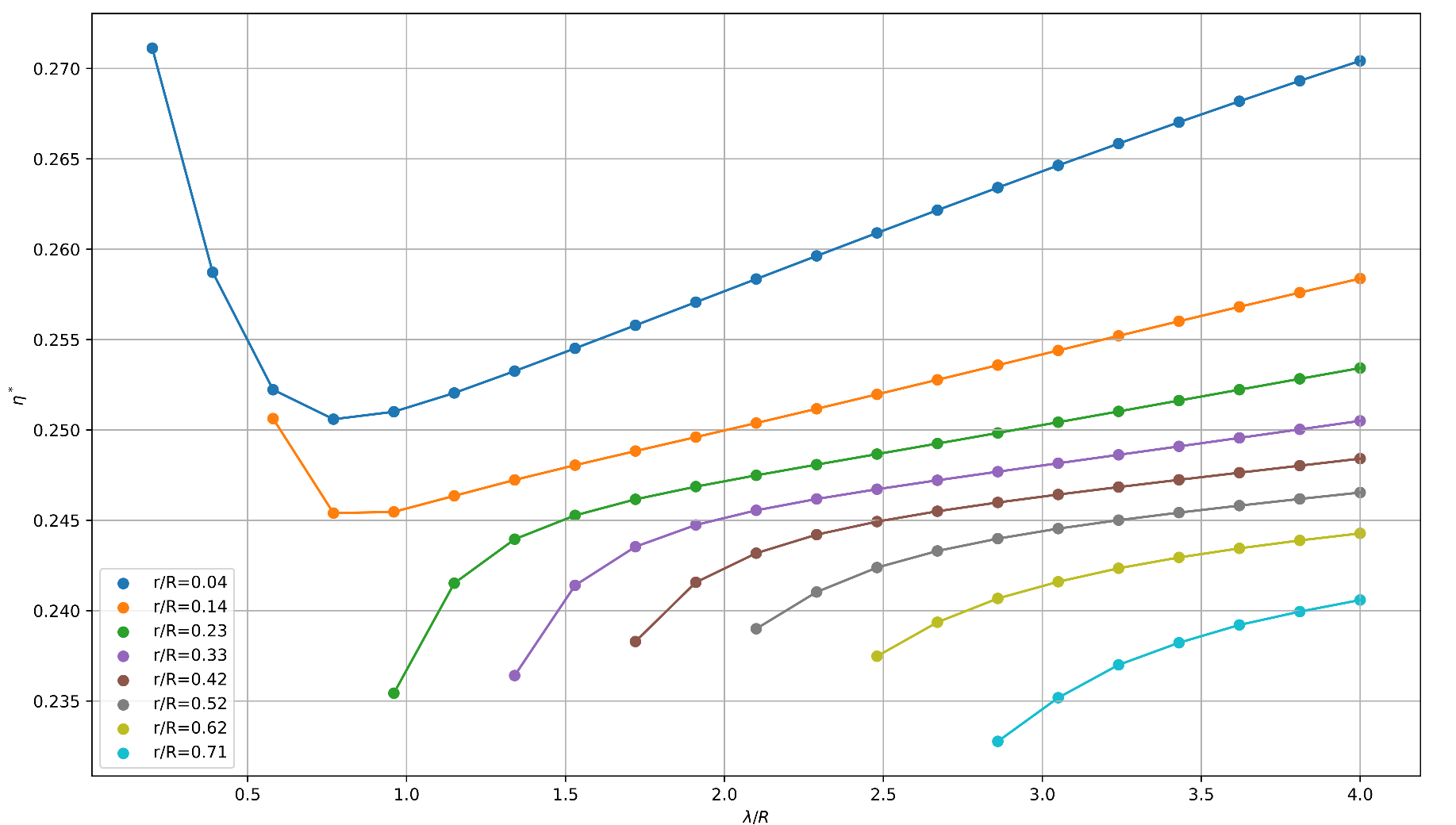}
\caption{Impact of $\lambda/R$ and $r/R$ on $\eta^*$(desired $\eta$ to have the same velocity for the filament and the micro-robot)} 
\label{combination_intersection}
\end{figure}

We now present results from simulations used to determine numerical values for the coefficients. The total length of the micro-robot was fixed at $L=1$, while the length of the filament was fixed at $L_f = 0.5$. Our analytical model does not explicitly account for the thickness of the filament. However, to allow a straight filament to fit inside the micro-robot, the ratio $r/R$ is constrained to take values between 0 and $(1-r_f/R)<1$. We consider pitches $\lambda > 4r$ so that successive turns of the helix are not too close together. In Figure \ref{combination_intersection}, we show $\eta^*$ for various values of $r/R$ between 0.04 and 0.71, and for $\lambda/R$ up to 4. Except for the smallest values of $r/R$ and $\lambda/R$, we find that $\eta^*$ increases with $\lambda/R$. In addition, we tested larger values of $\lambda/R$ up to 400 for $r/R=0.3$. Across all tested parameter combinations, the values of $\eta^*$ remained in a narrow range, between 0.23 and 0.275. This suggests that $\eta \approx 0.25$ should be suitable for most helical geometries. For comparison, the optimum value for $\xi$ was found to be 0.52 in Section \ref{sec:three_helix}, corresponding to $\eta = 0.234$.

In Section~\ref{sec:three_helix}, we mentioned two design considerations for the three-section micro-robot. To explain why the speed of the micro-robot increases with the total length, note that $\eta^*$, given by \eqref{eq:eta_star}, is a decreasing function of $L/L_f$ if as anticipated, $L>L_f$, $\beta < 0$, $0<\alpha < 1$, and $A_{12}/A_{11}<0$. Additionally, the swimming speed, given by \eqref{eq:robot_velocity}, is a decreasing function of $\eta$. Hence, the swimming speed for micro-robots designed with $\eta = \eta^*$ increases with the total length $L$.

The second consideration was that the bottom of the filament should be within the middle section of the micro-robot. To illustrate the benefit of this, suppose that the filament is straight and consider a vertical perturbation $\Delta Z$ of the micro-robot relative to the filament. If $\Delta Z < 0$ (the filament protrudes out of the micro-robot slightly), then the portion of the filament in the middle, pushing, section of the micro-robot is shorter while the portion in the front, pulling, section is the same. Hence, the filament will be pulled back into the micro-robot. If $\Delta Z > 0$, then the portion of the filament in the pulling helices becomes shorter while the portion in the pushing helix is longer. The filament is pushed forward and $\Delta Z$ decreases. Hence, this design stabilizes the fully coupled configuration $\Delta Z = 0$, as demonstrated in Figure~\ref{FullyCoupling}.

\section{Conclusions}
Recent advancements in micro-robot design have led to the development of robots capable of carrying cargo in vivo/in-vitro using a magnetic fields. However, many of these methods involve contact-based cargo transportation, which can make it difficult to release the cargo after reaching their target. In this study, we sought to provide a non-contact method for carrying cargo using a rotating helical micro-robot.

To design the helical micro-robot for capturing and carrying a passive filament, we first studied the streamline and pathline of fluid around the rotating helical micro-robot. To achieve a good coupling between the passive filament and the micro-robot, the filament must be located where the fluid flows in the forward direction. Our calculations showed that confining the filament inside the micro-robot close to its axis maximized the forward hydrodynamic drag force. We found that the geometrical parameters $r/R$ and $\lambda/R$ affect the pathline of the fluid significantly. Unsuitable choices for $r/R$ and $\lambda/R$ cause the pathline of the fluid to escape the radial confinement of the micro-robot, which according to the streamline of the fluid, the velocity of the fluid is in the backward direction.

Next, we simulated the motion of the helical micro-robot and the passive filament and examined the effects of three parameters: the flexibility of the filament, $r/R$, and $\lambda/R$. Our results indicated that increasing the flexibility of the filament and the ratio of $r/R$  led to better coupling, while reducing $\lambda/R$ improved the coupling as well. However, we found that the forward velocity of the filament for the designed helical micro-robot was lower than that of the micro-robot, causing the micro-robot to surpass the filament and lose the coupling. To address this issue, we added a pushing helix to the front of the micro-robot to apply a larger hydrodynamic thrust force to the filament and enhance its forward velocity.

Despite this improvement, we encountered another issue with the second design: the flow field on the front of the micro-robot was in the forward direction, causing the filament to push forward during the capturing process and making it difficult to capture. Thus, we proposed a third design, which consisted of three helical sections: one pushing microhelix located between two pulling microhelices of different lengths. The pulling helix located at the back provided the main thrust for the micro-robot to push forward, while the front helix pulled fluid into the micro-robot to assist with capturing the filament. By changing the length ratio between the pushing and pulling microhelices and selecting the helix's parameters of $r/R$ and $\lambda/R$, we were able to manipulate the coupling between the filament and the micro-robot. Using the optimal length ratio for coupling, our simulations demonstrated that the micro-robot with three sections was also effective at capturing a filament. 

Finally, we developed a theoretical model to estimate the length ratios that allow complete coupling between the filament and micro-robot and found a low sensitivity to geometrical parameters. The optimal length ratio from the model was in close agreement with the numerically determined optimum. In addition, the model supports our recommendation that the combined length of the front and middle sections be slightly longer than the length of the intended filament cargo. 

We anticipate that the micro-robot we have designed would be suitable for transporting immotile sperm cells. If the helical radius is large enough for the head to fit through, then the sperm can be carried without relying on direct contact with the micro-robot to push the cell forward. This could reduce the chance of adhesion and damage due to shear stresses applied to the head.

\section*{Acknowledgements}
We acknowledge the support of the Natural Sciences and Engineering Research Council of Canada (NSERC), [funding reference number RGPIN-2018-04418].

Cette recherche a \'{e}t\'{e} financ\'{e}e par le Conseil de recherches en sciences naturelles et en g\'{e}nie du Canada (CRSNG), [num\'{e}ro de r\'{e}f\'{e}rence RGPIN-2018-04418].

%%%%%%%%%% Insert bibliography here %%%%%%%%%%%%%%

%nocite{*}
\bibliographystyle{unsrt}

\end{document}